\documentclass[lettersize,journal]{IEEEtran}
\usepackage{amsmath,amsfonts}
\usepackage{algorithmic}
\usepackage{algorithm}
\usepackage{array}
\usepackage[caption=false,font=normalsize,labelfont=sf,textfont=sf]{subfig}
\usepackage{textcomp}
\usepackage{stfloats}
\usepackage{url}
\usepackage{verbatim}
\usepackage{graphicx}
\usepackage{cite}
\usepackage{booktabs}
\usepackage{makecell}
\usepackage{multirow}
\usepackage{colortbl}
\usepackage[dvipsnames]{xcolor}
\begin{document}

\title{VFM-Loc: Training-Free Cross-View Geo-Localization via Aligning Discriminative Visual Hierarchies}

\author{Jun Lu,
        Zehao Sang,
        Haoqi Wei,
        Xiangyun Liu,
        Kun Zhu,
        Haitao Guo,
        Zhihui Gong,
        and Lei Ding%
\thanks{Jun Lu, Zehao Sang, Haoqi Wei, Xiangyun Liu, Kun Zhu, Haitao Guo, Zhihui Gong, and Lei Ding are with Information Engineering University, Zhengzhou, China.}
\thanks{Lei Ding is also with Chinese Academy of Sciences, Beijing, China.}
\thanks{Corresponding author: Lei Ding.(e-mail:dinglei14@outlook.com)}}

\markboth{Manuscript Under Review.}%
{Shell \MakeLowercase{\textit{et al.}}: A Sample Article Using IEEEtran.cls for IEEE Journals}

\IEEEpubid{0000--0000/00\$00.00~\copyright~2021 IEEE}

\maketitle

\begin{abstract}
Cross-View Geo-Localization (CVGL) in remote sensing aims to locate a drone-view query by matching it to geo-tagged satellite images. Although supervised methods have achieved strong results on close-set benchmarks, they often fail to generalize to unconstrained, real-world scenarios due to severe viewpoint differences and dataset bias. To overcome these limitations, we present VFM-Loc, a training-free CVGL framework that leverages the generalizable visual representations from vision foundational models (VFMs). VFM-Loc identifies and matches discriminative visual clues across different viewpoints through a progressive alignment strategy. First, we design a hierarchical clue extraction mechanism using Generalized Mean pooling and Scale-Weighted R-MAC to preserve distinctive visual clues across scales while maintaining hierarchical confidence. Second, we introduce a statistical manifold alignment pipeline based on domain-wise PCA and Orthogonal Procrustes analysis, linearly aligning heterogeneous feature distributions in a shared metric space. Experiments demonstrate that VFM-Loc exhibits high accuracy on standard benchmarks and surpasses supervised methods by over 20\% in Recall@1 on the challenging LO-UCV dataset with large oblique angles. This work highlights that principled alignment of pre-trained features can effectively bridge the cross-view gap, establishing a robust and training-free paradigm for real-world CVGL. The relevant code is made available at: \url{github.com/DingLei14/VFM-Loc}.
\end{abstract}

\begin{IEEEkeywords}
Cross-View Geo-Localization, Remote Sensing, Training-Free Adaptation, Vision Foundation Models
\end{IEEEkeywords}

\section{Introduction}
\IEEEPARstart{C}{ross}-View Geo-Localization (CVGL) has emerged as a pivotal technology in autonomous drone navigation and urban Remote Sensing (RS). By establishing a spatial correspondence between oblique UAV images and nadir-view satellite maps, CVGL enables precise positioning in environments where traditional geolocation systems are unavailable or unreliable. Over the past decade, the rapid development of Deep Learning (DL) has advanced the accuracy of CVGL on standard benchmarks to near saturation. For instance, on widely used UAV-satellite CVGL benchmarks such as University-1652 and SUES-200, state-of-the-art (SOTA) approaches have achieved an average precision exceeding 95\%. However, such high accuracy is largely confined to closed-set evaluation scenarios. The real-world deployment of CVGL still faces significant generalization bottleneck. Existing supervised methods rely heavily on large volumes of paired UAV-satellite data, often leading to overfitting to specific sensor characteristics or geographic distributions. Consequently, these models suffer substantial accuracy degradation when applied to unseen regions or under different imaging conditions, thereby limiting their utility in real-world applications.

The rise of Visual Foundation Models (VFMs), pretrained in meta-scale data, offers a promising pathway toward learning universal visual representations. DINOv3 \cite{sim2025dinov3}, for example, has demonstrated an remarkable capability in acquiring universal visual representations with strong generalization. Nevertheless, we observe that a naive "plug-and-play" application of VFM features does not yield satisfactory localization accuracy. For instance, using the plain DINOv3 features with average pooling results in a Recall@1 of only 21.56\% in cross-view retrieval. This accuracy gap stems from the inherent distribution shift bewteen UAV and satellite viewpoints. Specifically, drone-view images, characterized by oblique viewing geometry and varying altitudes, exhibits fundamentally different spatial and semantic layouts compared to nadir-view satellite images. This disparity leads to a severe geometric and coverage mismatch, where a comprehensive drone scene may correspond to only a localized patch within a satellite tile. Furthermore, unconstrained environmental clutter and redundant information in real-world scenes further hinder effective matching.\IEEEpubid{}

\begin{figure}[t]
\centering
    \includegraphics[width=1\linewidth]{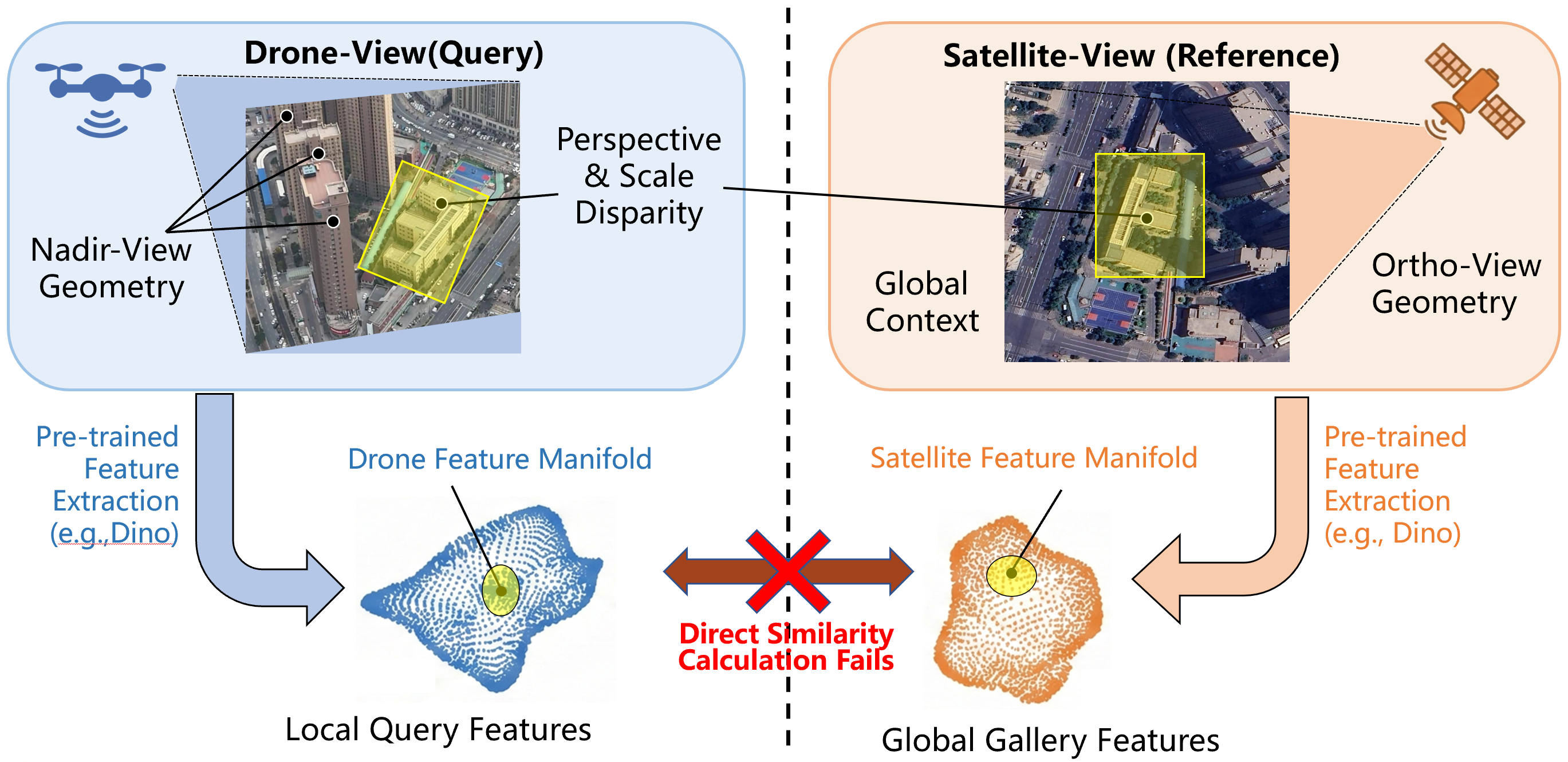}
    \caption{Key challenges in training-free CVGL: disparity in feature distribution. Drone and satellite visual manifolds suffer severe misalignment due to inherently different viewing geometry, thus direct similarity calculation fails under viewpoint discrepancies.}
\label{fig.flow_chart}
\end{figure}

To bridge the cross-view gap without exhaustive fine-tuning, this paper presents a novel training-free framework for UAV-satellite CVGL. Instead of learning a view-transformation mapping, our approach directly aligns the feature manifold and enhances discriminative saliency. The core contributions are threefold:

\begin{enumerate}

    \item A Training-free CVGL Paradigm. We present a localization framework built on a key insight: cross-view discrepancy in VFMs is primarily statistical and can be aligned linearly, without learning complex transforms. By leveraging pre-trained semantic priors embeded in VFMs, our method achieves high localization accuracy while substantially improving adaptability across diverse real-world scenarios.
    \item Manifold Alignment via Orthogona Transformation. To address the distribution shift across viewpoints, we introduce a domain-wise alignment strategy. Using Orthogonal Procrustes analysis, we derive an optimal linear transformation that aligns the heterogeneous feature distributions into a common metric space, effectively improving the retrieval accuracy.
    \item Multi-Granularity Saliency-Aware Similarity Modeling. We design a hierarchical feature aggregation scheme that integrates Generalized Mean (GeM) pooling with scale-weighted R-MAC. It explicitly captures salient landmarks at multiple spatial granularities, ensuring robust similarity computation under substantial imaging variation and background interference.
\end{enumerate}

\section{Related Work}

\subsection{Cross-View Geo-Localization}
CVGL aims to geo-localize query drone-captured image by matching it against a database of geo-referenced satellite imagery, fundamentally framing it as an image retrieval problem across heterogeneous viewpoints \cite{vo2016localizing, zhu2021vigor}. Advances in DL-based feature representation have been a primary driver of progress in this field \cite{durgam2024cross}. Early DL approaches commonly employed Siamese networks \cite{tian2017cross} or metric learning objectives \cite{liu2019lending} to learn embeddings that minimize the distance between correctly matched image pairs.

To bridge the severe appearance gap induced by extreme viewpoint differences, later work explored more sophisticated architectures and strategies. A key direction has been the use of polar transformation to geometrically align satellite/aerial images with ground-level panoramas \cite{liu2019lending, shi2019spatial}, significantly easing the learning problem and leading to substantial performance gains. Concurrently, research has evolved from relying solely on global features to incorporating local and structural information. This includes methods that explore multiscale consistency with enhanced feature interaction \cite{yang2026ecsnet}, employ optimal transport to model spatial correspondences \cite{shi2020optimal}, and design explicit part-based or local pattern networks to capture fine-grained details \cite{wang2021each, tian2021uav, wuqiong2024camp}. The introduction of vision transformers marked another leap, leveraging their strength in modeling long-range dependencies to enhance feature representation and extract robust geometry-aware features \cite{yang2021cross, zhu2022transgeo}. 
Meanwhile, vision-language models have been explored to bridge cross-view gaps via text-image guided feature extraction \cite{cao2026vlgeo}. Recent frameworks have moved beyond one-to-one retrieval by integrating joint retrieval-and-calibration pipelines that account for positional offsets, enabling finer-grained localization \cite{wang2023fine}. Furthermore, emerging generative methods, such as generative adversarial networks and diffusion models, are being investigated to synthesize cross-view imagery \cite{ye2025leveraging}, offering a promising path to alleviate data scarcity and viewpoint-gap challenges in CVGL.

Despite these advances, CVGL faces persistent bottleneck problems in cross-domain generalization. A central issue arises from the simplified assumptions underlying many benchmark datasets. Existing benchmarks often assume perfect center alignment between ground and aerial image pairs \cite{yang2021cross}, an assumption that is over-simplified in practical applications. Real-world scenarios frequently involve large positional offsets, which substantially degrade global feature similarity, introduces disruptive elements, and demands robust partial matching capabilities that many methods lack \cite{xia2025cross}. Consequently, generalizing CVGL to unseen geographic regions and handling unconstrained viewpoint and positional discrepancies remains an open and critical challenge.

\subsection{Unsupervised Learning}
The evolution of unsupervised learning, particularly moving toward training-free paradigms, has transformed computer vision from early methods like simple clustering and autoencoders to modern self-supervised frameworks centered on contrastive learning and masked image modeling (MIM). Classic contrastive approaches, such as MoCo \cite{he2020momentum}, leverage momentum updates and queue mechanisms to learn highly discriminative feature representations from unlabeled data. A recent survey \cite{gui2024survey} highlights that MIM-based methods like MAE further enhance contextual semantic understanding by reconstructing masked regions. A clear trend is the growing shift toward training-free adaptation \cite{pourpanah2022review}, which utilizes large-scale pretrained vision-language models to generalize to unseen categories \cite{zhou2023zegclip}.Recent literature indicates that the research focus has expanded beyond pure image-feature alignment to leveraging natural language descriptions and multimodal foundation models for open-vocabulary understanding and reasoning \cite{wu2025logiczsl, ye2025cross}. This shift aims to overcome the limitations of traditional models confined to closed-set training and evaluation.

Unsupervised CVGL face significant challenges due to the severe viewpoint discrepancy. In \cite{regmi2018cross} conditional GANs are employed for cross-view image synthesis to bridge the domain gap. OriCNN \cite{liu2019lending} introduced orientation as a geometric prior for unsupervised feature alignment. Recent research has increasingly explored leveraging large-scale unlabeled data. For instance, in \cite{li2024unleashing} an unsupervised paradigm is  proposed to addresses the cold-start problem via pseudo-labeling and correspondence-free projection, enabling robust localization without GPS labels. Concurrently, cluster-based approach first generate intra-view pseudo-labels via clustering and dual-contrastive learning, then align representations across views using cosine similarity \cite{Chen2025UniABGUA}. Another direction focuses on adapter-based methods that use vision foundation models as backbones, incorporating lightweight adapters to project VFM features into a uniform space, training them via asymmetric contrastive learning on pseudo-labeled samples \cite{li2024learning}. However, achieving truly unsupervised training-free CVGL remains a formidable challenge. While existing methods reduce the dependency on annotated data, they still require domain-specific finetuning with labels. Meanwhile, VFMs still struggle with extreme geometric deformations. Current solutions usually depend on domain-specific unsupervised adaptation or few-shot fine-tuning to align feature spaces, which limits their ability to achieve fully plug-and-play, training-free localization on unseen data.

\section{Methodology}

\subsection{Problem Formulation and System Overview}
\label{sec:blind}

\begin{figure*}[t]
\centering
    \includegraphics[width=0.8\linewidth]{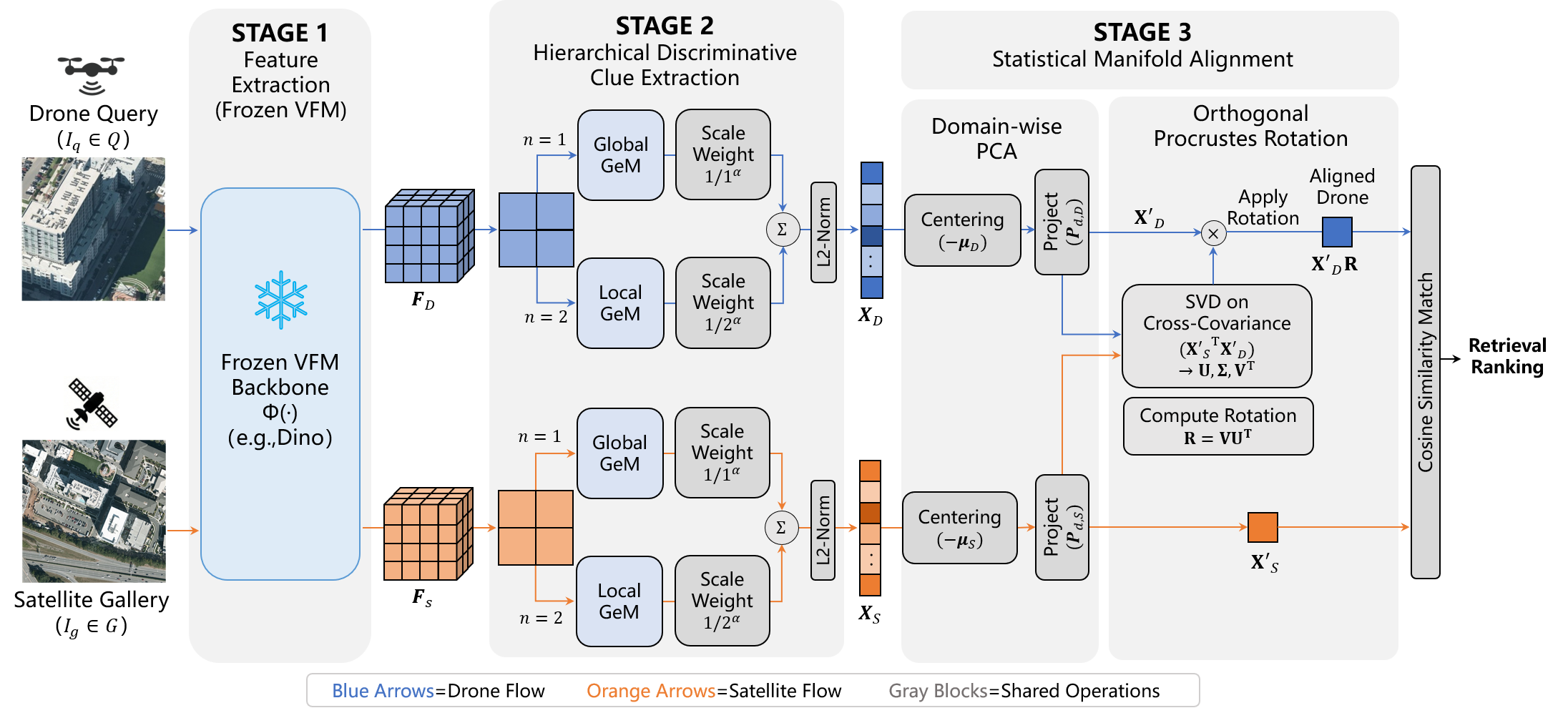}
    \caption{The proposed VFM-Loc, a training-free CVGL framework, operates in three training-free stages: (1) Feature extraction using a VFM; (2) Hierarchical discriminative clue extraction via GeM pooling and scale-weighted R-MAC; (3) Statistical manifold alignment based on domain-wise PCA and orthogonal Procrustes rotation, projecting heterogeneous features into a unified embedding space for retrieval.}
\label{fig.flow_chart}
\end{figure*}

In training-free CVGL, we aim to retrieve the correct satellite image $I_g \in \mathcal{G}$ given a drone query $I_q \in \mathcal{Q}$, with no task-specific tuning. As shown in Fig.~\ref{fig.flow_chart}, we adopt a frozen VFM encoder $\Phi(\cdot)$ to map images into a $D$-dimensional feature space. $I_q, I_g$ are then encoded into global feature vectors $\mathbf{z}_d, \mathbf{z}_s$ via pooling operations. The core challenge stems from distributional shifts between the drone domain $\mathcal{D}$ and the satellite domain $\mathcal{S}$. Due to drastic viewpoint difference (oblique vs. nadir), the learned representations of a same region exhibit heterogeneity in the VFM space, i.e., $P(\mathbf{z}|\mathcal{D}) \neq P(\mathbf{z}|\mathcal{S})$. This gap appears not merely as a feature-space translation, but as an anisotropic manifold distortion. Consequently, direct similarity metrics such as Euclidean distance $\|\mathbf{z}_d - \mathbf{z}_s\|_2$ are thus dominated by domain variance rather than geographic identity.

Our proposed VFM-Loc framework addresses this challenge through three sequential stages without any backpropagation. First, as described, frozen VFM features are extracted. Second, a statistical manifold alignment pipeline applies domain-wise PCA followed by Orthogonal Procrustes (OP) analysis. This step finds an optimal linear transformation $\mathcal{T}$ that minimizes cross-domain divergence while preserving semantic topology, effectively aligning $\mathcal{T}(\mathbf{z}_d)$ and $\mathbf{z}_s$ in a common metric space. Third, we refine the alignment by integrating a hierarchical scheme that combines GeM pooling \cite{dollar2009integral} with scale-weighted Regional Maximum Activation of Convolutions (R-MAC). This multi-granularity fusion accentuates discriminative landmark regions and suppresses background clutter, resulting in robust aggregated features for similarity ranking.

\subsection{Hierarchical Extraction of Discriminative Visual Clues}
A critical prerequisite for alignment is the extraction of robust descriptors that are invariant to environmental noise yet sensitive to geometric landmarks. Standard Global Average Pooling (GAP) and Global Max Pooling (GMP) represent two extremes of information aggregation. GAP ($p=1$) treats all spatial activations equally, which tends to filter discriminative landmarks with non-informative background noise. Conversely, GMP ($p \to \infty$) only considers the single peak activation, making it highly susceptible to transient outliers or specular reflections. To achieve a superior balance, we employ GeM Pooling, defined for a feature map $\mathbf{X} \in \mathbb{R}^{C \times H \times W}$ as:

\begin{equation}
    \mathbf{f} = [f_1, \dots, f_k, \dots, f_C]^\top, \quad f_k = \left( \frac{1}{N} \sum_{i=1}^N x_{k,i}^p \right)^{1/p}
\end{equation}
where $p$ is the pooling parameter. Mathematically, GeM acts as a differentiable soft-maximum. As $p$ increases, the pooling operation assigns exponentially greater weight to higher-intensity activations, which in the context of VFMs, typically correspond to discriminative visual clues such as building corners or road intersections. By setting $p > 1$, GeM effectively suppresses the influence of low-activation regions (e.g., grass, shadows) while avoiding the over-fitting to single pixels seen in GMP. This focused design ensures that the resulting descriptor captures essential structural cues of a scene, providing a more stable input for the subsequent manifold alignment.

Building upon the GeM-refined features, we further address the global-local mismatch by introducing a R-MAC \cite{tolias2015particular} strategy. As the UAV perspective often captures only a portion of the landmarks present in a satellite tile, we partition the feature map into multi-scale grids ($n \times n$) to encode spatial structures at varying granularities. To suppress local noise and retain global context, a scale-dependent decay factor $1/n^\alpha$ weights the contribution of each level, yielding the aggregated descriptor:

\begin{equation}
    \mathbf{V}_{agg} = \text{Norm} ( \sum_{n} \frac{1}{n^\alpha} \text{GeM}(\mathbf{F}_n) )
\end{equation}
where $\alpha$ is a decay factor that controls the confidence of local vs. global clues.

This formulation prioritizes the global context (where the VFM's generalization reliability is highest) while allowing local discriminative clues to refine the matching score. The resulting hierarchical descriptor $\mathbf{V}_{agg}$ ensures that spatial discrepancies do not lead to catastrophic matching failures, significantly enhancing the framework's robustness in open-set environments.

\subsection{Statistical Manifold Alignment}

Despite the robustness of extracted clues, the generalization performance of VFMs is frequently hampered by domain shift in the feature space. We attribute this to a misalignment in the statistical distribution of cross-view features, despite their shared underlying semantics. To rectify this, we introduce a dual-stage alignment pipeline comprising domain-wise PCA Pre-conditioning and OP rotation.

\textbf{Domain-wise PCA Pre-conditioning:} This stage is to discard redundant feature dimensions and suppress non-semantic noise within the raw VFM output. We first apply Principal Component Analysis (PCA) independently to the query and gallery manifolds.
Let $\mathbf{X} \in \mathbb{R}^{N \times D}$ be the feature extracted from a specific domain (Drone or Satellite), we first compute the domain-specific mean vector:

$$\boldsymbol{\mu} = \frac{1}{N} \sum_{i=1}^N \mathbf{x}_i$$

The first-order alignment is then achieved by shifting the disparate manifolds to a common origin through a zero-centering operation:

$$\bar{\mathbf{X}} = \mathbf{X} - \mathbf{1}\boldsymbol{\mu}^\top$$

We subsequently derive the domain-specific covariance matrix:

$$\mathbf{C}_{pca} = \frac{1}{N-1} \bar{\mathbf{X}}^\top \bar{\mathbf{X}}$$

and perform eigen-decomposition such that:

$$\mathbf{C}_{pca} = \mathbf{P} \mathbf{\Lambda} \mathbf{P}^\top$$
where $\mathbf{P}$ is the matrix of eigenvectors and $\mathbf{\Lambda}$ is the diagonal matrix of eigenvalues. By selecting the top $d$ eigenvectors $\mathbf{P}_d \in \mathbb{R}^{D \times d}$ that capture the majority of the variance, we project the centered features into a de-noised subspace:

$$\mathbf{X}' = \bar{\mathbf{X}} \mathbf{P}_d$$

This projection isolates the most significant spatial-semantic variations while discarding low-variance components typically associated with environmental noise. By projecting the features onto their primary principal components, we isolate the most significant spatial-semantic variations, thereby preparing the disparate manifolds for a more refined linear alignment. 

\textbf{Orthogonal Procrustes Alignment:} Following PCA-based centering, we apply Orthogonal Procrustes (OP) alignment \cite{schonemann1966generalized} to transform the drone and satellite feature manifolds into a common metric space. Given the centered features $\mathbf{X}_D$ (Drone) and $\mathbf{X}_S$ (Satellite), we solve for the optimal orthogonal transformation $\mathbf{R}$ between the two domains:

\begin{equation}
    \min_{\mathbf{R}} \|\mathbf{X}_D \mathbf{R} - \mathbf{X}_S\|_F^2, \quad \text{s.t. } \mathbf{R}^\top \mathbf{R} = \mathbf{I}
\end{equation}

The solution is obtained via Singular Value Decomposition (SVD) of the cross-domain covariance matrix:

\begin{equation}
    \mathbf{U}\mathbf{\Sigma}\mathbf{V}^\top = \text{SVD}(\mathbf{X}_S^\top \mathbf{X}_D)
\end{equation}
where $\mathbf{U}$ and $\mathbf{V}$ are orthogonal matrices representing the principal axes of the satellite and drone manifolds, and $\mathbf{\Sigma}$ contains the singular values representing the strength of these correlations. This decomposition distills the principal shared mapping directions between the two heterogeneous domains. To enforce strict orthogonality (i.e., $\mathbf{R}^\top \mathbf{R} = \mathbf{I}$), we discard the scaling component $\mathbf{\Sigma}$ and recombine the singular vectors to derive an operator consisting purely of rotation:
\begin{equation}
    \mathbf{R} = \mathbf{V}\mathbf{U}^\top
\end{equation}

This alignment preserves the intrinsic semantic topology of the VFM space while matching the statistical distributions across domains. Consequently, the drone-view manifold is robustly transformed toward the satellite-view manifold with training-free computation, efficiently bridging the viewpoint gap and substantially enhancing retrieval accuracy.

\section{Experiments}

\subsection{CVGL Benchmarks and Evaluation Metrics}

We evaluate the proposed VFM-Loc on three CVGL benchmarks to comprehensively assess its performance and generalizability. The standard University-1652 dataset is used to establish baseline and compare against SOTA methods. We further employ SUES-200, which introduces complexity through varied UAV flight altitudes, to test robustness under diverse observational conditions. To rigorously evaluate real-world generalization, we introduce a new dataset, the Large-Oblique UAV-Satellite Cross-View (LO-UCV) dataset. This dataset features greater scene diversity and significant oblique viewing angles, presenting a more challenging and practical testing scenario. Detailed specifications of all datasets are provided in Table \ref{tab:datasets}.

\begin{figure*}[t]
\centering
    \includegraphics[width=1\linewidth]{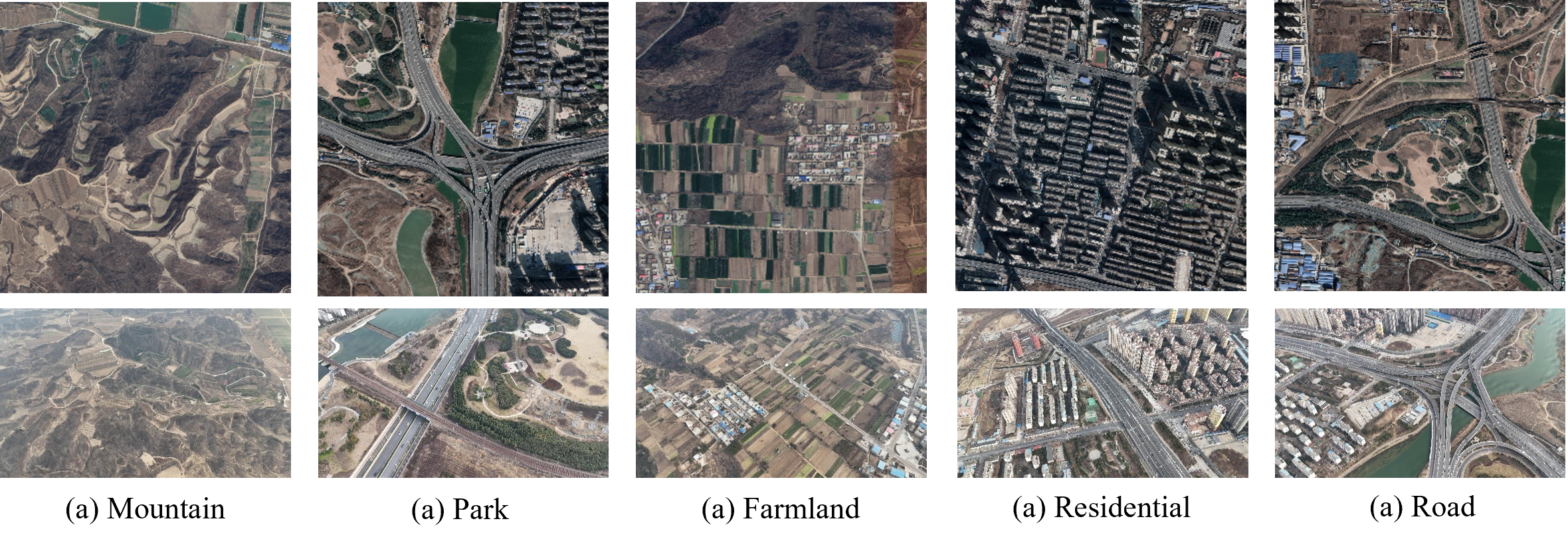}
  
    \caption{Sample image pairs from the constructed LO-UCV dataset. Top: satellite images; Bottom: high-resolution images from fixed-wing UAVs with large tilt ($>45^\circ$).}
    \label{fig:dataset_examples}
\label{fig.vis_sim}
\end{figure*}

We specifically developed the LO-UCV dataset to overcome the common issues of scene monotony and limited viewpoint variation in existing CVGL datasets. Utilizing a DJI Mini4 Pro UAV flying at a fixed altitude of 500 meters over Zhengzhou with an oblique shooting angle exceeding 45°, the dataset captures 3,407 high-resolution (3840×2160) images. These images encompass a wide range of RS scene types, including roads, villages, urban areas, farmland, rivers, and parks, across distinct geographic locations, ensuring both diversity and representativeness. Corresponding high-quality satellite images were obtained from the Google Earth platform based on the UAV’s GPS coordinates, and manual verification ensured precise geo-referencing while eliminating redundant frames. The dataset is split into training and test sets with strict separation to prevent any scene overlap: the training set comprises 42 satellite images paired with 2,235 drone images, and the test set comprises 22 satellite images paired with 1,172 drone images.  Figure \ref{fig:dataset_examples} illustrates typical challenging scenarios from the LO-UCV dataset, showcasing its rich viewpoint variations and complex scene content. This dataset provides a robust test benchmark for evaluating the generalization capability and practical applicability of CVGL algorithms.

Following the standard CVGL evaluation protocol \cite{Deuser2023Sample4GeoHN}, we employ Recall@1 and Average Precision (AP) as our primary metrics. Recall@1 reports the rate at which the true match is retrieved as the top result, reflecting localization precision under the strictest condition. AP, defined as the area under the precision-recall curve, evaluates the overall retrieval quality across all recall levels, offering a more comprehensive and threshold-robust assessment.

\begin{table*}[t]
\caption{Statistical Summary of the Experimental Datasets}
\label{tab:datasets}
\centering
\scriptsize
\setlength{\tabcolsep}{3pt}
\resizebox{\textwidth}{!}{%
\begin{tabular}{llllll}
\toprule
Datasets & Platform & GSD / Resolution & Image Size & Data Size & Key Characteristics \\
\midrule
University-1652 [33] & UAV+Sat & Varies & 512$\times$512 & 146,580 images & Multi-view, multi-source benchmark \\
SUES-200 [35] & UAV+Sat & 0.13 m--0.92 m & 512$\times$512 & 40,200 images & Four distinct UAV flight altitudes \\
LO-UCV (constructed) & UAV+Sat & UAV: 0.2 m; Sat: 0.6 m & UAV: 340$\times$2160; Sat: 768$\times$768 & 3,471 images & Large oblique angle ($>45^\circ$), diverse scenes \\
\bottomrule
\end{tabular}%
}
\end{table*}

\subsection{Ablation Study}

\textbf{Effectiveness of the proposed modules.} We conduct a systematic ablation study to validate each component of VFM-Loc. As reported in Table \ref{tab:ablation_incremental_full}, the vanilla DINOv3 features yield only 21.56\% R@1 (drone→satellite), underscoring the challenge posed by the cross-view domain gap. Replacing average pooling with class token pooling, a classification mechanism provided in Dinov3, offers a slight gain (+4.84\% R@1), but remains far from satisfactory for practical usage.

The introduction of our SMA module brings a dramatic improvement. By aligning the two domains via PCA and OP, SMA boosts the best R@1 to 73.30\%, an absolute gain of +51.74\% over the baseline. This dramatic increase indicates that the primary obstacle for training-free cross-view localization is not a lack of semantic discriminability in the foundation model, but the statistical discrepancy between the two viewing domains. SMA effectively aligns the feature manifolds of drone and satellite imagery into a common metric space, enabling robust retrieval without any task-specific training.

The integration of GeM pooling and R‑MAC further refines the representation. The complete pipeline (SMA + GeM + R-MAC) achieves 74.42\% R@1 on drone→satellite and 90.44\% R@1 on satellite→drone retrieval, surpassing the baseline by +52.86\% and +76.89\%, respectively. The results confirm that SMA alleviates the fundamental domain gap, while GeM and R-MAC consistently enhance feature saliency, collectively enabling accurate training-free CVGL.

\begin{table*}[t]
\caption{Ablation study on University-1652.}
\label{tab:ablation_incremental_full}
\centering
\scriptsize
\setlength{\tabcolsep}{4pt}
\renewcommand{\arraystretch}{1.08}
\resizebox{\textwidth}{!}{%
\begin{tabular}{lcccccccccc}
\toprule
\multirow{2}{*}{\textbf{Methods}}
& \multicolumn{5}{c}{Drone $\rightarrow$ Satellite} 
& \multicolumn{5}{c}{Satellite $\rightarrow$ Drone} \\
\cmidrule(lr){2-6} \cmidrule(lr){7-11}
& R@1 & R@5 & R@10 & R@top1 & AP 
& R@1 & R@5 & R@10 & R@top1 & AP \\
\midrule

\textbf{Baseline:} DINOv3 (+avg pooling)
& 21.56 & 40.82 & 50.83 & 49.20 & 31.38
& 13.55 & 24.25 & 30.39 & 83.88 & 9.57 \\
\midrule

\rowcolor[HTML]{EFEFEF} \multicolumn{11}{l}{\textbf{A. Replace pooling strategies}} \\
\quad avg $\rightarrow$ cls
& 26.40 & 47.45 & 57.16 & 55.65 & 36.79
& 23.54 & 36.23 & 42.08 & 83.02 & 13.13 \\
\quad avg $\rightarrow$ GeM
& 7.04 & 20.57 & 28.92 & 27.50 & 14.55
& 18.40 & 30.53 & 40.94 & 92.44 & 11.79 \\
\quad \textit{Best $\Delta$ vs. baseline}
& \textcolor{ForestGreen}{$\uparrow$ 4.84} & \textcolor{ForestGreen}{$\uparrow$ 6.63} & \textcolor{ForestGreen}{$\uparrow$ 6.33} & \textcolor{ForestGreen}{$\uparrow$ 6.45} & \textcolor{ForestGreen}{$\uparrow$ 5.41}
& \textcolor{ForestGreen}{$\uparrow$ 10.00} & \textcolor{ForestGreen}{$\uparrow$ 12.00} & \textcolor{ForestGreen}{$\uparrow$ 11.69} & \textcolor{BrickRed}{$\downarrow$ 0.86} & \textcolor{ForestGreen}{$\uparrow$ 3.56} \\
\midrule

\rowcolor[HTML]{FFE6E6} \multicolumn{11}{l}{\textbf{B. Apply PCA directly} \textit{(failed attempt)}} \\
\quad DINOv3 (avg) + PCA
& 0.00 & 0.00 & 0.29 & 1.00 & 0.08
& 0.70 & 2.62 & 4.55 & 4.17 & 2.59 \\
\quad \textit{$\Delta$ vs. baseline}
    & \textcolor{BrickRed}{$\downarrow$ 21.56} & \textcolor{BrickRed}{$\downarrow$ 40.82} & \textcolor{BrickRed}{$\downarrow$ 50.54} & \textcolor{BrickRed}{$\downarrow$ 48.20} & \textcolor{BrickRed}{$\downarrow$ 31.30}
    & \textcolor{BrickRed}{$\downarrow$ 12.85} & \textcolor{BrickRed}{$\downarrow$ 21.63} & \textcolor{BrickRed}{$\downarrow$ 25.84} & \textcolor{BrickRed}{$\downarrow$ 79.71} & \textcolor{BrickRed}{$\downarrow$ 6.98} \\
\midrule

\rowcolor[HTML]{E6F2FF} \multicolumn{11}{l}{\textbf{C. SMA} \textit{(PCA + OP)}} \\
\quad DINOv3 (avg) + PCA + OP
& 68.74 & 91.39 & 95.35 & 94.91 & 78.78
& 88.73 & 97.43 & 98.86 & 99.86 & 67.84 \\
\quad DINOv3 (cls) + PCA + OP
& 63.72 & 88.83 & 93.66 & 93.06 & 74.88
& 83.17 & 95.86 & 98.00 & 100.00 & 61.13 \\
\quad DINOv3 (GeM) + PCA + OP
& 73.30 & 93.77 & 96.96 & 96.58 & 82.42
& 90.73 & 98.15 & 99.14 & 100.00 & 72.33 \\
\quad \textit{Best $\Delta$ vs. baseline}
& \textcolor{ForestGreen}{$\uparrow$ 51.74} & \textcolor{ForestGreen}{$\uparrow$ 52.95} & \textcolor{ForestGreen}{$\uparrow$ 46.13} & \textcolor{ForestGreen}{$\uparrow$ 47.38} & \textcolor{ForestGreen}{$\uparrow$ 51.04}
& \textcolor{ForestGreen}{$\uparrow$ 77.18} & \textcolor{ForestGreen}{$\uparrow$ 73.90} & \textcolor{ForestGreen}{$\uparrow$ 68.75} & \textcolor{ForestGreen}{$\uparrow$ 16.12} & \textcolor{ForestGreen}{$\uparrow$ 62.76} \\
\midrule

\rowcolor[HTML]{E6F2FF} \multicolumn{11}{l}{\textbf{D. VFM-Loc} \textit{(PCA + OP + R-MAC)}} \\
\quad DINOv3 (avg) + PCA + OP + R-MAC
& 68.25 & 91.03 & 95.18 & 94.68 & 78.36
& 88.16 & 97.29 & 98.57 & 99.86 & 67.16 \\
\quad DINOv3 (cls) + PCA + OP + R-MAC
& 68.27 & 91.02 & 95.19 & 94.68 & 78.35
& 88.16 & 97.29 & 98.57 & 99.86 & 67.16 \\
\quad DINOv3 (GeM) + PCA + OP + R-MAC
& \textbf{74.42} & \textbf{94.16} & \textbf{97.17} & \textbf{96.91} & \textbf{83.18}
& \textbf{90.44} & \textbf{98.43} & \textbf{98.86} & \textbf{100.00} & \textbf{72.41} \\
\quad \textit{Best $\Delta$ vs. baseline}
& \textcolor{ForestGreen}{$\uparrow$ 52.86} & \textcolor{ForestGreen}{$\uparrow$ 53.34} & \textcolor{ForestGreen}{$\uparrow$ 46.34} & \textcolor{ForestGreen}{$\uparrow$ 47.71} & \textcolor{ForestGreen}{$\uparrow$ 51.80}
& \textcolor{ForestGreen}{$\uparrow$ 76.89} & \textcolor{ForestGreen}{$\uparrow$ 74.18} & \textcolor{ForestGreen}{$\uparrow$ 68.47} & \textcolor{ForestGreen}{$\uparrow$ 16.12} & \textcolor{ForestGreen}{$\uparrow$ 62.84} \\
\bottomrule
\end{tabular}%
}
\end{table*}

\begin{figure*}[!t]
\centering
{%
    \includegraphics[width=0.75\linewidth]{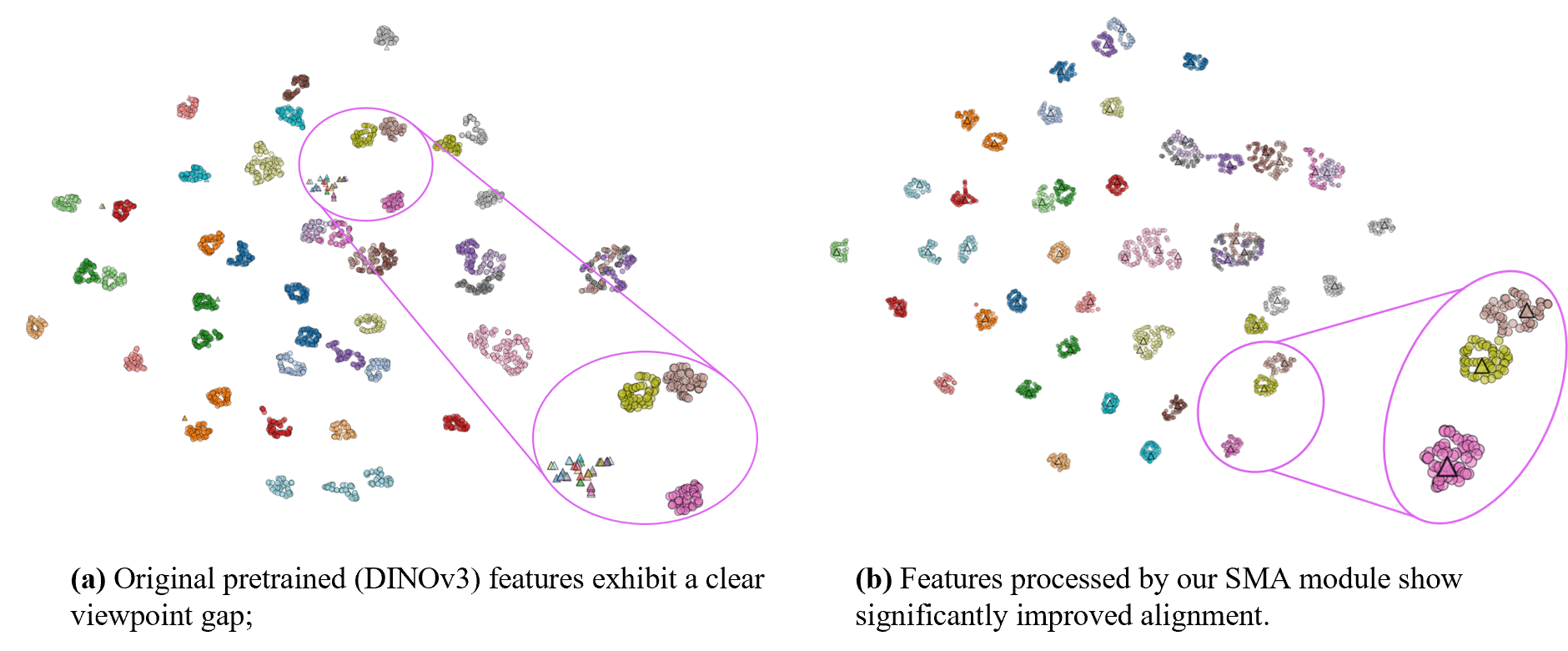}

}
\caption{T-SNE visualization of features before (a) and after (b) SMA alignment. Circles: UAV features; triangles: satellite features, colored by different regions. In (b), cross-view features cluster closely.}
    \label{fig:t-SNE}
\end{figure*}

\textbf{Visualization on Manifold Alignment.} To validate the impact of the SMA module, we further visualize the cross-view feature distributions using t-SNE. Colors indicate different geographic locations sampled randomly from the dataset; shapes denote the viewpoint (triangles represent satellite-view features, and circles represent UAV-view features). As shown in Fig.~\ref{fig:t-SNE}(a), before applying SMA, feature clusters are clearly separated by viewpoint: samples of the same color (i.e., same location) are split into distinct groups, with satellite triangles appearing far from their corresponding UAV circles. This reveals a pronounced domain gap between the two domains. After applying SMA (Fig.~\ref{fig:t-SNE}(b)), satellite features migrate toward the centers of their respective color clusters, aligning closely with corresponding UAV samples. This visual result confirms that SMA successfully brings heterogeneous cross-view distributions into a unified metric space, a key factor for robust cross-view retrieval.

\begin{figure}[t]
\centering
    \includegraphics[width=1\linewidth]{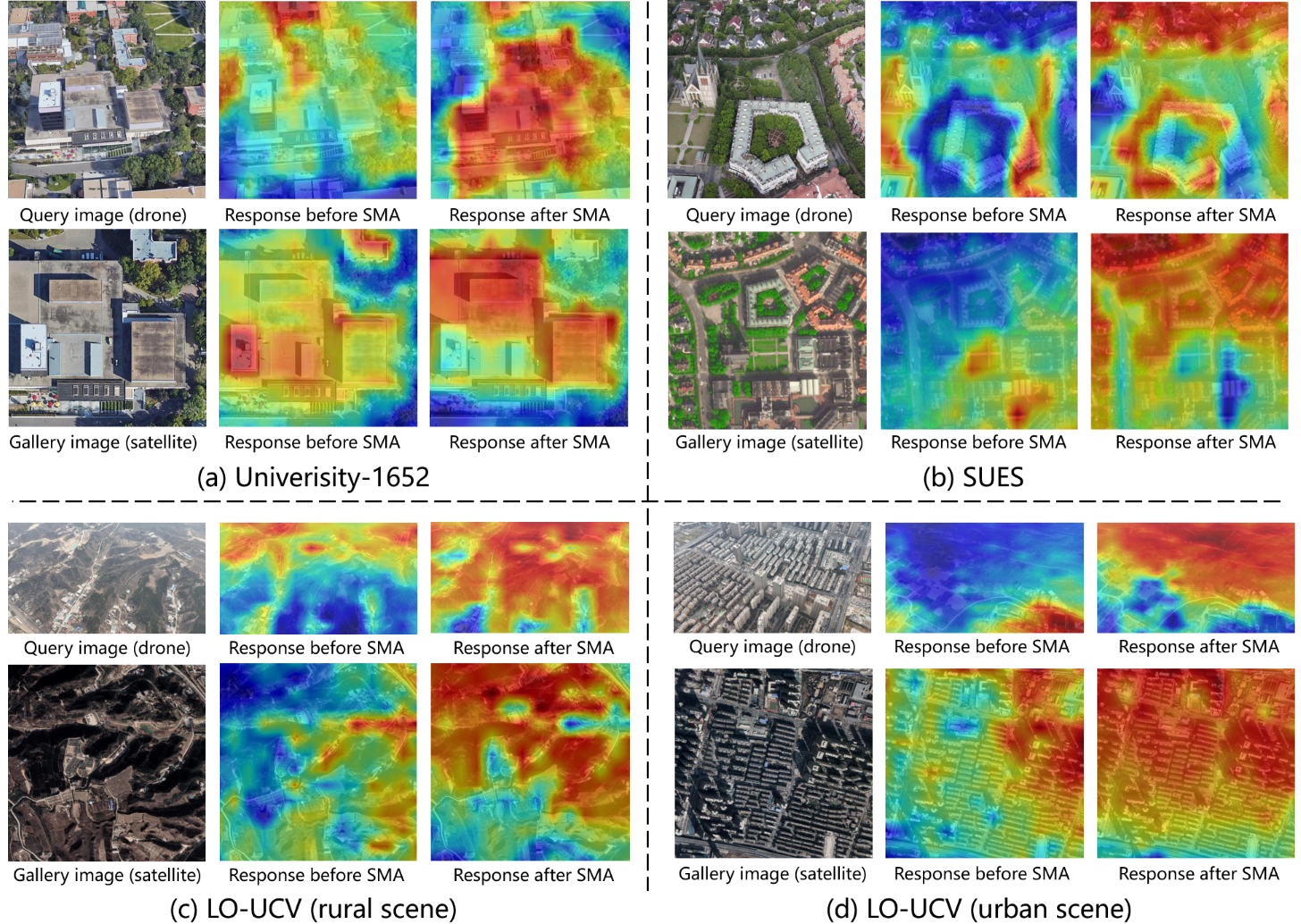}
    \caption{Similarity response visualization before and after manifold alignment.}
\label{fig.vis_sim}
\end{figure}

\textbf{Visualization of Similarity Responses.} To qualitatively evaluate the proposed manifold alignment module (e.g., SMA), we visualize spatial similarity distributions of typical UAV-satellite image pairs before and after alignment. We extract patch-level features from DINOv3, retaining their spatial layout. After performing domain-wise PCA and Orthogonal Prostrustes analysis, the resulting rotation matrix is applied to the UAV patch features. For each spatial location, we compute cosine similarity between the UAV features and the global pooled satellite feature, then normalize and upsample to generate heatmaps.

The visualization in Fig.~\ref{fig.vis_sim} reveals that SMA significantly mitigates the cross-view distribution shift present in raw VFM features. Before alignment, similarity heatmaps are often diffuse and activated by background clutter, reflecting the geometric and semantic gap between oblique drone and nadir satellite viewpoints. After alignment, the feature responses become highly concentrated and consistently localize geometrically stable structures, such as distinctive roof outlines and road intersections, even under substantial viewpoint variation. This indicates that the learned orthogonal transformation successfully projects the heterogeneous feature manifolds into a shared metric space, suppressing view-specific noise while enhancing discriminative cross-view landmarks. The results validate that our training-free approach with principled manifold alignment yields reliable local matching cues, which underpin the strong generalization performance observed in our quantitative experiments.

\setcounter{table}{3}
\begin{table*}[!t]
  \centering
  \caption{Accuracy of the VFM-Loc using different VFMs (University-1652).}
  \label{tab:model_comparison}
  \setlength{\tabcolsep}{5pt}
  \resizebox{\textwidth}{!}{%
  \begin{tabular}{l c c c c c c c c c c c c}
    \toprule
    \multirow{2}{*}{VFMs}
    & \multirow{2}{*}{Params (M)}
    & \multirow{2}{*}{FLOPs (G)}
    & \multicolumn{5}{c}{Drone $\rightarrow$ Satellite}
    & \multicolumn{5}{c}{Satellite $\rightarrow$ Drone} \\
    \cmidrule(lr){4-8} \cmidrule(lr){9-13}
    & &
    & R@1 & R@5 & R@10 & R@top1 & AP
    & R@1 & R@5 & R@10 & R@top1 & AP \\
    \midrule
    SAM3 \cite{carion2025sam3}
    & 838.21 & 2514.58
    & 55.49 & 78.32 & 84.88 & 81.60 & 60.69
    & 66.90 & 83.74 & 87.59 & 98.15 & 30.94 \\
    DINOv3 \cite{sim2025dinov3}
    & \textbf{85.67} & \textbf{12.93}
    & 74.42 & 94.16 & 97.17 & 96.91 & 83.18
    & 90.44 & \textbf{98.43} & 98.86 & 100.00 & 72.41 \\
    RadioV4-H \cite{Ranzinger2026CRADIOv4}
    & 651.65 & 129.60
    & \textbf{76.36} & \textbf{94.90} & \textbf{97.55} & \textbf{97.24} & \textbf{84.53}
    & \textbf{92.58} & 98.00 & \textbf{99.43} & \textbf{100.00} & \textbf{75.86} \\
    RadioV4-S0400M \cite{Ranzinger2026CRADIOv4}
    & 431.24 & 84.68
    & 72.02 & 92.48 & 96.24 & 95.74 & 81.08
    & 90.01 & \textbf{98.43} & 99.14 & 100.00 & 71.43 \\
    \bottomrule
  \end{tabular}%
  }
\end{table*}

\setcounter{table}{2}
\begin{table}[t]
\caption{Ablation study on the decay factor $\alpha$ used in R-MAC.}
\label{tab:alpha}
\centering
\scriptsize
\setlength{\tabcolsep}{3pt}
\resizebox{\columnwidth}{!}{%
\begin{tabular}{lcccccccccc}
\toprule
\multirow{2}{*}{$\alpha$} & \multicolumn{5}{c}{Drone $\rightarrow$ Satellite} & \multicolumn{5}{c}{Satellite $\rightarrow$ Drone} \\
\cmidrule(lr){2-6} \cmidrule(lr){7-11}
& R@1 & R@5 & R@10 & R@top1 & AP & R@1 & R@5 & R@10 & R@top1 & AP \\
\midrule
1 & 73.29 & 93.77 & 96.98 & 96.63 & 82.36 & 89.30 & 98.15 & 98.86 & 100.00 & 71.80 \\
2 & 73.80 & 93.95 & 97.13 & 96.75 & 82.74 & 89.73 & 98.00 & 98.86 & 100.00 & 72.01 \\
3 & 74.12 & 94.07 & 97.23 & 96.85 & 82.98 & 90.16 & 98.15 & 98.86 & 100.00 & 72.20 \\
4 & 74.32 & 94.14 & 97.21 & 96.89 & 83.12 & \textbf{90.58} & 98.43 & 98.86 & 100.00 & 72.33 \\
5 & 74.38 & \textbf{94.18} & \textbf{97.18} & 96.89 & 83.16 & 90.16 & 98.43 & 98.86 & 100.00 & 72.39 \\
6 & \textbf{74.42} & 94.16 & 97.17 & \textbf{96.91} & \textbf{83.18} & 90.44 & 98.43 & \textbf{98.86} & \textbf{100.00} & 72.41 \\
7 & 74.42 & 94.18 & 97.17 & 96.89 & 83.18 & 90.58 & 98.43 & 98.86 & 100.00 & 72.42 \\
8 & 74.40 & 94.17 & 97.16 & 96.89 & 83.18 & 90.58 & \textbf{98.57} & 98.86 & 100.00 & \textbf{72.43} \\
9 & 74.42 & 94.17 & 97.16 & 96.88 & 83.18 & 90.58 & 98.57 & 98.86 & 100.00 & 72.43 \\
\bottomrule
\end{tabular}%
}
\end{table}

\setcounter{table}{4}

\textbf{Impact of $\alpha$ in R-MAC.} We further analyze the sensitivity of the R-MAC decay factor $\alpha$, which balances global context against local discriminative clues. As shown in Table \ref{tab:alpha}, performance remains consistently high across a wide range of $\alpha$. An initial rise in $\alpha$ shifts the weighting toward more stable global structures, improving results. The best and most stable performance is achieved when $\alpha$ lies between 6 and 9, confirming the stability of the hierarchical weighting scheme. We select $\alpha=6$ as our default setting, as it consistently delivers near-best results across both retrieval tasks.

\textbf{Effect of Different VFMs.} We evaluate four representative VFMs, including SAM3 \cite{carion2025sam3}, DINOv3\cite{sim2025dinov3} and Radio~\cite{heinrich2025radiov25, Ranzinger2026CRADIOv4}, to analyze how their inherent designs affect the CVGL performance. Four representative VFMs are evaluated to analyze how their pre-training objectives and architecture design affect zero‑shot CVGL performance: SAM3 \cite{carion2025sam3}, DINOv3 \cite{sim2025dinov3}, RadioV4‑huge, and RadioV4‑s0400 \cite{Ranzinger2026CRADIOv4}. As summarized in Table \ref{tab:model_comparison}, VFMs pre-trained primarily for dense prediction (e.g., the segmentation-oriented SAM3) perform poorly due to their limited discriminative embedding ability. In contrast, models optimized for generic representation learning through self-supervised (DINOv3) or large-scale contrastive objectives (RadioV4) achieve substantially higher accuracy. Among them, the larger RadioV4-huge delivers the best overall results, highlighting the positive impact of model scale and pre-training data volume.

An important trade-off emerges when considering computational efficiency. While RadioV4-huge attains top accuracy, it entails significantly higher parameters and FLOPs compared to DINOv3 ($>5$ times in computation costs). DINOv3 provides a highly favorable efficiency-accuracy balance: its performance is only marginally lower than that of RadioV4-huge, yet it is dramatically more efficient. Therefore, considering the trade-off for practical deployment, we adopt DINOv3 as the primary backbone for all subsequent experiments.

\subsection{Comparative Experiments}


\textbf{Quantitative Comparisons with SOTA.} To thoroughly evaluate our approach, we compare it with leading supervised and unsupervised methods on three CVGL benchmarks. Supervised baselines include PCL\cite{9583266}, Sample4Geo\cite{Deuser2023Sample4GeoHN} and DAC\cite{10636268}; unsupervised baselines are EM-CVGL\cite{li2024learning} and UniABG\cite{Chen2025UniABGUA}. It is worth noting that both necessitate training on the target dataset, while our method operates in a fully training-free manner.

\begin{table*}[!t]
\caption{Comparison with SOTA methods on the University-1652 dataset (R@1\% and AP\%).}
\label{tab:university1652_comparison}
\centering
\footnotesize
\setlength{\tabcolsep}{4pt}
\resizebox{0.75\linewidth}{!}{%
\begin{tabular}{cclcccccc}
\toprule
\multirow{2}{*}{Label-Free} & \multirow{2}{*}{Training-Free} & \multirow{2}{*}{\textbf{Method}} 
& \multicolumn{2}{c}{\textbf{Drone $\rightarrow$ Sat}} & \multicolumn{2}{c}{\textbf{Sat $\rightarrow$ Drone}} 
& \multicolumn{2}{c}{\textbf{Inference Cost}} \\
\cmidrule(lr){4-5} \cmidrule(lr){6-7} \cmidrule(lr){8-9}
& & & R@1 & AP & R@1 & AP & Params (M) & FLOPs (G) \\
\midrule
\multirow{5}{*}{$\times$}  & \multirow{5}{*}{$\times$} 
& University-1652 \cite{zheng2020university} & 59.69 & 64.80 & 73.18 & 59.40 & 52.52 & 24.29 \\
& & PCL \cite{9583266} & 79.47 & 83.63 & 87.69 & 78.51 & - & - \\
& & Sample4Geo \cite{Deuser2023Sample4GeoHN} & 92.65 & 93.81 & 95.14 & 91.39 & 84.57 & 90.24 \\
& & DAC \cite{10636268} & 94.67 & 95.50 & 96.43 & 93.79 & 96.50 & 90.24 \\
& & QDFL \cite{10955695} & \textbf{95.00} & \textbf{95.83} & \textbf{97.15} & \textbf{94.57} & 108.87 & 80.24 \\
\midrule
\multirow{3}{*}{$\checkmark$}  & \multirow{3}{*}{$\times$}
& EM-CVGL \cite{li2024learning} & 70.82 & 75.36 & 79.03 & 61.03 & 1140+6.30 & 372.19+18.92 \\
& & Wang et al. \cite{FromCoarsetoFine}  & 85.95 & 90.33 & 94.01 & 82.66 & - & - \\
& & UniABG \cite{Chen2025UniABGUA} 
& 93.62 & 94.61 & 95.43 & 93.29 & 96.50 & 45.12 \\
\midrule
\multirow{2}{*}{$\checkmark$} & \multirow{2}{*}{$\checkmark$} 
& VFM-Loc (DINOv3) 
74.42 & 83.18 & 90.44 & 72.41 & \textbf{85.67} & \textbf{12.93} \\
& & VFM-Loc (RadioV4) 
& 76.36 & 84.53 & 92.58 & 75.86 & 431.24 & 84.68 \\
\bottomrule
\end{tabular}%
}
\end{table*}

As shown in Table~\ref{tab:university1652_comparison}, our VFM-Loc (RadioV4) achieves a competitive R@1 of 76.36\% for Drone→Satellite and 92.58\% for Satellite→Drone queries on the University-1652 dataset. Notably, this training-free method already outperforms several prior unsupervised approaches that require explicit training on the data, validating that a strong baseline can be established using only pre-trained VFMs. While recent unsupervised methods (e.g., UniABG) obtain higher accuracy, they necessitate computationally expensive training or fine-tuning on the target dataset. In contrast, our training-free framework demonstrates the effectiveness of manifold alignment and saliency modeling in adapting off-the-shelf VFM features to the CVGL task.

We further analyze the inference efficiency of each method as summarized in Table~\ref{tab:university1652_comparison}. The computational statistics for some baselines are absent, as their official implementations are unavailable. For EM-CVGL, a two-stage method, the reported figures correspond to the sum of both stages. Among methods with available data, VFM-Loc (DINOv3) achieves the lowest FLOPs while delivering competitive accuracy. Coupled with its training-free paradigm, which entirely eliminates training overhead, our VFM-Loc presents itself as the most computationally lightweight and readily deployable approach.

\begin{table*}[t]
\caption{Comparison of performance with SOTA methods on the SUES-200 dataset. S: Supervised; U: Unsupervised.}
\label{tab:sues_comparison}
\centering
\footnotesize
\setlength{\tabcolsep}{3pt}
\resizebox{\textwidth}{!}{%
\begin{tabular}{clcccccccccccccccc}
\toprule
\multirow{3}{*}{\textbf{Type}} 
& \multirow{3}{*}{\textbf{Method}} 
& \multicolumn{8}{c}{\textbf{Drone $\rightarrow$ Satellite}} 
& \multicolumn{8}{c}{\textbf{Satellite $\rightarrow$ Drone}} \\
\cmidrule(lr){3-10} \cmidrule(lr){11-18}
& 
& \multicolumn{2}{c}{150m} 
& \multicolumn{2}{c}{200m} 
& \multicolumn{2}{c}{250m} 
& \multicolumn{2}{c}{300m} 
& \multicolumn{2}{c}{150m} 
& \multicolumn{2}{c}{200m} 
& \multicolumn{2}{c}{250m} 
& \multicolumn{2}{c}{300m} \\
\cmidrule(lr){3-4} \cmidrule(lr){5-6} \cmidrule(lr){7-8} \cmidrule(lr){9-10}
\cmidrule(lr){11-12} \cmidrule(lr){13-14} \cmidrule(lr){15-16} \cmidrule(lr){17-18}
& & R@1 & AP & R@1 & AP & R@1 & AP & R@1 & AP & R@1 & AP & R@1 & AP & R@1 & AP & R@1 & AP \\
\midrule
\multirow{5}{*}{\textbf{S}} 
& SUES-200 \cite{zhu2023sues} 
& 55.65 & 61.92 & 66.78 & 71.55 & 72.00 & 76.43 & 74.05 & 78.26 
& 75.00 & 55.46 & 85.00 & 66.05 & 86.25 & 69.94 & 88.75 & 74.46 \\
& FSRA \cite{9648201} 
& 68.25 & 73.45 & 83.00 & 85.99 & 90.68 & 92.27 & 93.46 & 91.95 
& 83.75 & 76.67 & 90.00 & 85.34 & 93.75 & 90.17 & 95.00 & 92.03 \\
& MCCG \cite{10185134} 
& 82.22 & 85.47 & 89.38 & 91.41 & 93.82 & 95.04 & 95.07 & 96.20 
& 93.75 & 89.72 & 93.75 & 92.21 & 96.25 & 96.14 & 98.75 & 96.64 \\
& DAC \cite{10636268} 
& 96.80 & 97.54 & 97.48 & 97.97 & 98.20 & 98.62 & 97.58 & 98.14 
& 97.50 & 94.06 & 98.75 & 96.66 & 98.75 & 98.09 & 98.75 & 97.87 \\
& QDFL \cite{10955695} 
& 93.97 & 95.42 & 98.25 & 98.67 & 99.30 & \textbf{99.48} & 99.31 & \textbf{99.48} 
& 98.75 & 95.10 & 98.75 & 97.92 & \textbf{100.00} & 99.07 & \textbf{100.00} & 99.07 \\
\midrule
\multirow{3}{*}{\textbf{U}} 
& Wang et al. \cite{FromCoarsetoFine} 
& 76.90 & 84.95 & 87.88 & 92.60 & 92.98 & 95.66 & 95.10 & 96.92 
& 87.50 & 74.81 & 92.50 & 87.15 & 96.25 & 91.20 & 98.75 & 94.52 \\
& UniABG \cite{Chen2025UniABGUA} 
& 92.40 & 93.95 & 97.32 & 97.92 & 98.07 & 98.55 & 98.67 & 98.98 
& 98.75 & 91.54 & 98.75 & 97.06 & \textbf{100.00} & 98.32 & 98.75 & 97.58 \\
& VFM-Loc (ours) 
& \textbf{99.62} & \textbf{99.35} & \textbf{99.52} & \textbf{99.33} & \textbf{99.63} & 99.38 & \textbf{99.60} & 99.38 
& \textbf{98.75} & \textbf{99.22} & \textbf{98.75} & \textbf{99.32} & 98.75 & \textbf{99.36} & 98.75 & \textbf{99.36} \\
\bottomrule
\end{tabular}%
}
\end{table*}

On the more complex SUES-200 dataset, our method sets a new state-of-the-art. As detailed in Table~\ref{tab:sues_comparison}, for the Drone→Satellite task, VFM-Loc achieves R@1 scores consistently exceeding 99.5\% across all altitudes, significantly outperforming the best supervised baseline (QDFL). It delivers a performance gain of over 5.6\% at the most challenging 150m height. For the Satellite→Drone task, our method either matches or exceeds the top supervised method across all settings. This highlights its capability to bridge the cross-view gap in complex urban scenarios without any fine-tuning.

\begin{table}[t]
\caption{Comparison with SOTA methods on the constructed LO-UCV dataset. S: Supervised; U: Unsupervised.}
\label{tab:supervised_unsupervised_compare}
\centering
\footnotesize
\setlength{\tabcolsep}{4pt}
\resizebox{\columnwidth}{!}{%
\begin{tabular}{clcccc}
\toprule
\multirow{2}{*}{\textbf{Type}} 
& \multirow{2}{*}{\textbf{Method}} 
& \multicolumn{2}{c}{\textbf{Drone $\rightarrow$ Sat}} 
& \multicolumn{2}{c}{\textbf{Sat $\rightarrow$ Drone}} \\
\cmidrule(lr){3-4} \cmidrule(lr){5-6}
& & R@1 (\%) & AP (\%) & R@1 (\%) & AP (\%) \\
\midrule
\multirow{2}{*}{\textbf{S}} 
& Sample4Geo \cite{Deuser2023Sample4GeoHN} & 53.07 & 62.02 & 59.09 & 53.85 \\
& DAC \cite{10636268} & 67.83 & 74.01 & 63.63 & 58.65 \\
\midrule
\multirow{3}{*}{\textbf{U}} 
& EM-CVGL \cite{li2024learning} & 47.27 & 57.27 & 45.45 & 47.28 \\
& UniABG \cite{Chen2025UniABGUA} & 44.62 & 53.06 & 40.91 & 43.02 \\
& VFM-Loc (ours) & \textbf{89.84} & \textbf{94.63} & \textbf{95.31} & \textbf{91.17} \\
\bottomrule
\end{tabular}%
}
\end{table}

The most compelling advantage of our method is observed on the LO-UCV dataset, characterized by significant viewpoint differences and limited data. As presented in Table~\ref{tab:supervised_unsupervised_compare}, both supervised and unsupervised methods struggle on this benchmark. In contrast, our VFM-Loc achieves 89.84\% and 95.31\% R@1 for Drone→Sat and Sat→Drone tasks, respectively. This constitutes a substantial accuracy leap, surpassing the best supervised method by over 20\%$\sim$30\% in R@1 for the two query directions, underscoring superior generalization under data-scarce and geometrically difficult conditions.

\begin{table*}[!t]
\caption{Cross-dataset generalization capability evaluated on the LO-UCV dataset.}
\label{tab:cross_modality_results}
\centering
\footnotesize
\setlength{\tabcolsep}{4pt}
\resizebox{\textwidth}{!}{%
\begin{tabular}{l c ccccc ccccc}
\toprule
\multirow{2}{*}{\textbf{Methods}} 
& \multirow{2}{*}{\textbf{Training Dataset}} 
& \multicolumn{5}{c}{\textbf{Drone $\rightarrow$ Sat}} 
& \multicolumn{5}{c}{\textbf{Sat $\rightarrow$ Drone}} \\
\cmidrule(lr){3-7} \cmidrule(lr){8-12}
& & R@1 & R@5 & R@10 & R@top1 & AP & R@1 & R@5 & R@10 & R@top1 & AP \\
\midrule
Sample4Geo & University-1652 & 46.42 & 79.10 & 93.09 & 46.42 & 53.74 & 50.00 & 50.00 & 54.55 & 54.55 & 43.60 \\
DAC & University-1652 & 50.60 & 82.25 & 97.70 & 50.60 & 57.64 & 50.00 & 50.00 & 54.55 & 54.55 & \underline{50.54} \\
EM-CVGL & University-1652 & \underline{59.04} & \underline{97.01} & \underline{100.00} & \underline{59.04} & 66.77 & 50.00 & 68.18 & 68.18 & 68.18 & 49.39 \\
UniABG & University-1652 & 51.54 & 91.13 & 97.87 & 51.54 & 59.36 & \underline{54.55} & \underline{68.18} & \underline{68.18} & \underline{68.18} & 50.47 \\
\midrule
Sample4Geo & SUES-200 & 48.55 & 91.81 & 96.42 & 48.55 & 57.75 & 50.00 & 59.09 & 59.09 & 59.09 & 54.51 \\
DAC & SUES-200 & \underline{55.20} & 89.68 & 96.84 & \underline{55.20} & \underline{62.93} & \underline{63.64} & \underline{68.18} & \underline{68.18} & \underline{68.18} & \underline{55.70} \\
EM-CVGL & SUES-200 & 47.27 & \underline{91.98} & \underline{99.74} & 47.27 & 57.27 & 45.45 & 59.09 & 63.64 & 63.64 & 47.28 \\
UniABG & SUES-200 & 32.68 & 78.07 & 98.89 & 32.68 & 43.39 & 40.91 & 45.45 & 45.45 & 50.00 & 39.76 \\
\midrule
VFM-Loc (ours) & None & \textbf{89.84} & \textbf{99.97} & \textbf{100.00} & \textbf{89.84} & \textbf{94.63} & \textbf{95.31} & \textbf{98.44} & \textbf{98.44} & \textbf{98.44} & \textbf{91.17} \\
\bottomrule
\end{tabular}%
}
\end{table*}

\textbf{Evaluation of Cross-Dataset Generalization.} To evaluate cross-dataset transfer, we conduct a systematic generalization experiment. The compared methods are first trained on two public datasets, University-1652 and SUES-200, and then directly tested on the unseen LO-UCV dataset. This setup simulates real-world deployment of a trained model to a new geographic region with distinct terrain and lanc-cover characteristics.

Results under this protocol (Table~\ref{tab:cross_modality_results}) show that supervised and unsupervised methods achieve only moderate and varying performance. When transfer from University-1652 and tested on LO-UCV, unsupervised EM-CVGL performs best for Drone→Satellite, with Recall@1 of 59.04\%. For Satellite→Drone, UniABG achieves the top Recall@1 of 54.55\%. When trained on SUES-200, the supervised DAC method reaches the highest performance across both query directions, with Recall@1 scores of 55.20\% and 63.64\% respectively. Notably, unsupervised methods exhibit considerable inconsistency across training datasets.

In sharp contrast, our VFM-Loc requires \emph{no training} on any source dataset, yet it demonstrates exceptional generalization. It attains 89.84\% Recall@1 for Drone→Satellite and 95.31\% for Satellite→Drone. This represents a decisive lead over all trained baselines, with improvements exceeding +30\% in Recall@1. This result strongly validates the superior transferability and robustness of our approach, underscoring its practical potential for deployment in unseen geographic environments.

\begin{figure*}
    \centering
    \includegraphics[width=1\linewidth]{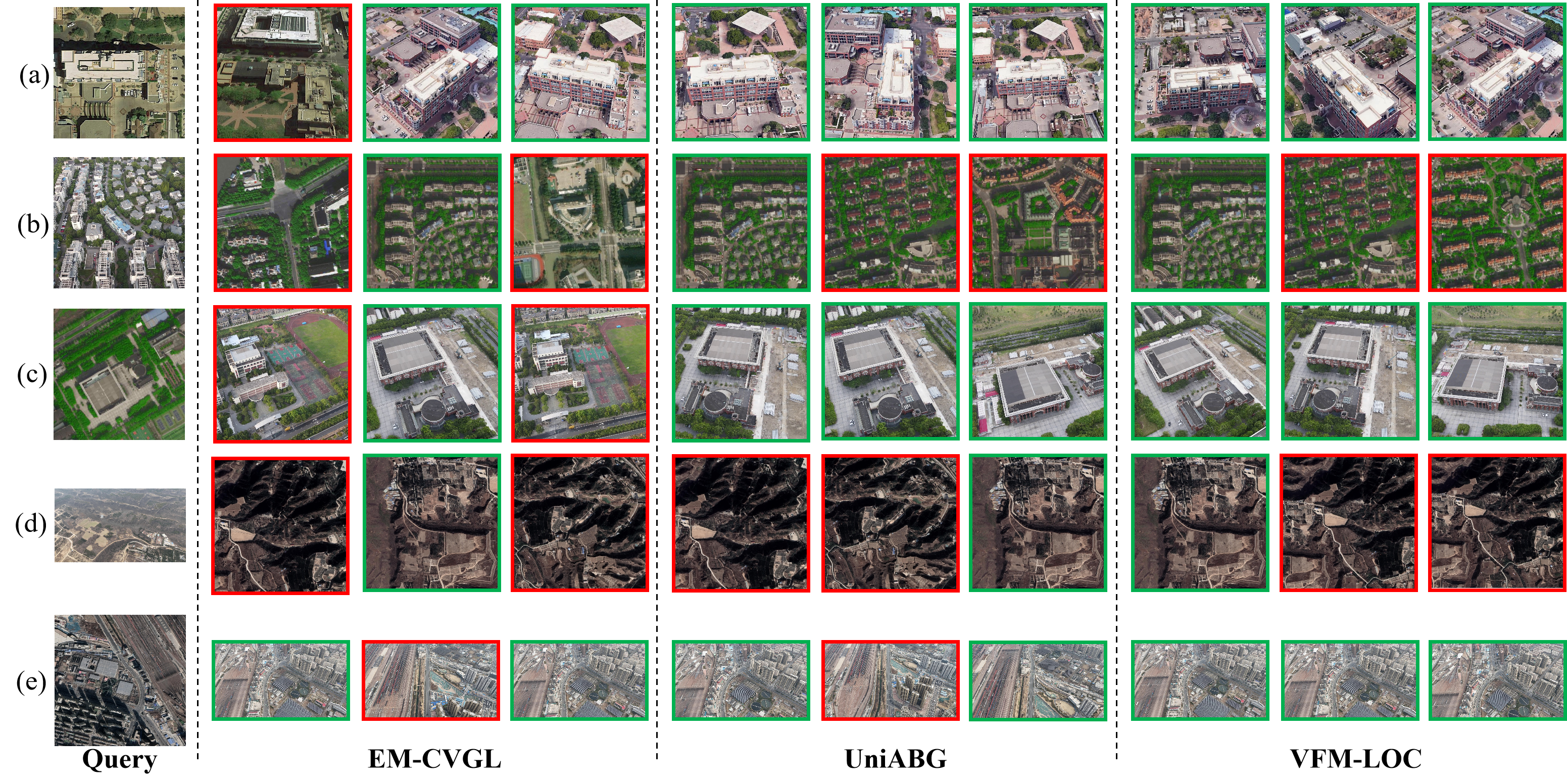}
    \caption{Comparison of qualitative results obtained on (a)University, (b)(c)SUES and (d)(e)LO-UCV.Green bounding boxes denote correct retrievals, while red bounding boxes denote incorrect retrievals.}
    \label{fig:retrieval_visualization}
\end{figure*}

\textbf{Qualitative Comparisons.} In Fig. \ref{fig:retrieval_visualization} we presents a qualitative comparison between our method and the SOTA unsupervised methods (EM-CVGL and UniABG). On the university, VFM-Loc correctly retrieves the satellite match for a drone query amid repetitive building facades, while EM-CVGL fails due to texture ambiguity. On the SUES, VFM-Loc distinguishes visually similar structures under occlusion and lighting variations, whereas unsupervised methods misalign due to shallow feature matching. On the challenging LO-UVB benchmark, VFM-Loc handles extreme viewpoint disparities by aligning hierarchical visual cues (e.g., landforms, street layouts), while competitors collapse to irrelevant retrieval. These results prove the robustness of our VFM-Loc to scene complexity, viewpoint gaps, and structural redundancy.
In summary, the comparative experiments across varied RS data reveal the distinct strengths of VFM-Loc. It provides a strong training-free baseline on standard benchmarks, delivers leading accuracy on complex multi-height urban scenarios, and demonstrates superior robustness under challenging, low-data regimes. These results confirm that by leveraging general visual features with a principled alignment strategy, VFM-Loc offers an advantageous balance of high performance, exceptional adaptability, and minimal data dependency, establishing a new practical paradigm for CVGL.

\section{Conclusion}
This work addresses the critical challenge of generalizing CVGL to unconstrained, real-world scenarios. We present VFM-Loc, a novel training-free framework that effectively bridges the severe viewpoint gap by aligning the discriminative visual hierarchies inherent in pre-trained VFMs. Our core contribution is a principled, two-stage strategy. First, we introduce a hierarchical clue extraction mechanism to generate robust multi-scale descriptors that emphasize stable geographical landmarks. Second, we develop a statistical manifold alignment pipeline that projects heterogeneous drone and satellite features into a unified, comparable metric space without any learning.

Extensive experiments on multiple benchmarks validate the effectiveness of our approach. VFM-Loc establishes a strong training-free baseline on standard benchmarks and demonstrates exceptional generalization. Notably, on the most challenging dataset with large oblique angles and limited data, our method achieves a performance gain exceeding 20\% in Recall@1 over the best supervised alternative. These results confirm that aligning pre-trained feature manifolds through a lightweight, data-efficient process can unlock the practical potential of VFMs for robust and scalable geo-localization, offering a promising paradigm for deployment in diverse and unseen environments.

%
%
\bibliographystyle{splncs04}
\bibliography{refs}
\end{document}